\documentclass[10pt,twocolumn,letterpaper]{article}

\usepackage{cvpr}              

\usepackage[dvipsnames]{xcolor}

\definecolor{cvprblue}{rgb}{0.21,0.49,0.74}
\usepackage[pagebackref,breaklinks,colorlinks,citecolor=cvprblue]{hyperref}
\usepackage{color,array}

\usepackage{graphicx}
\usepackage{amsmath}
\usepackage{amssymb}
\usepackage{booktabs}
\usepackage{pifont}
\usepackage{tikz}
\newcommand{\tikzxmark}{%
\tikz[scale=0.23] {
    \draw[line width=0.7,line cap=round] (0,0) to [bend left=6] (1,1);
    \draw[line width=0.7,line cap=round] (0.2,0.95) to [bend right=3] (0.8,0.05);
}}
\newcommand{\tikzcmark}{%
\tikz[scale=0.23] {
    \draw[line width=0.7,line cap=round] (0.25,0) to [bend left=10] (1,1);
    \draw[line width=0.8,line cap=round] (0,0.35) to [bend right=1] (0.23,0);
}}

\usepackage{comment}
\usepackage{adjustbox}
\usepackage{tabularx}
\usepackage{xcolor}
\newcolumntype{a}{>{\columncolor{lightmintbg}}c}
\usepackage{makecell}
\usepackage{color, colortbl}
\definecolor{Gray}{gray}{0.9}
\definecolor{LightCyan}{rgb}{0.88,1,1}
\definecolor{maroon}{cmyk}{0,0.87,0.68,0.32}
\definecolor{babyblueeyes}{rgb}{0.63, 0.79, 0.95}
\definecolor{beaublue}{rgb}{0.5, 0.5, 0.7}
\definecolor{mintbg}{rgb}{.63,.79,.95}
\colorlet{lightmintbg}{mintbg!30}
\newcolumntype{L}{>{\raggedright\arraybackslash}c}
\usepackage[linesnumbered,ruled]{algorithm2e}
\usepackage{multirow}
\SetKwInput{KwInput}{Input}                
\SetKwInput{KwOutput}{Output}              
\SetKwInOut{KwInit}{Initialization}
\def\paperID{13080} 
\def\confName{CVPR}
\def\confYear{2024}

\title{Unknown Sample Discovery for Source Free Open Set Domain Adaptation}
 
\author{Chowdhury Sadman Jahan\\
Center for Imaging Science\\
Rochester Institute of Technology, Rochester, NY, USA\\
{\tt\small sj4654@rit.edu}
\and
Andreas Savakis\\
Department of Computer Engineering\\
Rochester Institute of Technology, Rochester, NY, USA\\
{\tt\small axseec@rit.edu}
}

\begin{document}
\maketitle
\begin{abstract}
Open Set Domain Adaptation (OSDA) aims to adapt a model trained on a source domain to a target domain that undergoes distribution shift and contains samples from novel classes outside the source domain. 
Source-free OSDA (SF-OSDA) techniques eliminate the need to access source domain samples, but current SF-OSDA methods utilize only the known classes in the target domain for adaptation, and require access to the entire target domain even during inference after adaptation, to make the distinction between known and unknown samples. 
In this paper, we introduce \textbf{U}nknown \textbf{S}ample \textbf{D}iscovery (USD) as an SF-OSDA method that utilizes a temporally ensembled teacher model to conduct known-unknown target sample separation and adapts the student model to the target domain over all classes using co-training and temporal consistency between the teacher and the student. 
USD promotes Jensen-Shannon distance (JSD) as an effective measure for known-unknown sample separation. 
Our teacher-student framework significantly reduces error accumulation resulting from imperfect known-unknown sample separation, while curriculum guidance helps to reliably learn the distinction between target known and target unknown subspaces. 
USD appends the target model with an unknown class node, thus readily classifying a target sample into any of the known or unknown classes in subsequent post-adaptation inference stages.
Empirical results show that USD is superior to existing  SF-OSDA methods and is competitive with current OSDA models that utilize both source and target domains during adaptation.
\end{abstract}    
\section{Introduction}
\label{sec:intro}

The domain gap manifests when a model, trained on a source domain with annotated samples, is deployed in a target domain that has a distribution shift compared to the source domain. 
Unsupervised domain adaptation (UDA) considers  a target domain with unlabeled data and aims to mitigate the domain gap by aligning the source and target domain distributions using the knowledge learned from the source domain. 
Most of the current UDA methods conduct domain adaptation by either minimizing the source-target distribution discrepancy \cite{torralba2011unbiased, zellinger2017central}, or by adversarially aligning the feature spaces of the source and target data \cite{ganin2016domain, hoffman2018cycada}. 
However, such models need access to the source data during adaptation, and therefore cannot be applied to cases where the source domain is not available or when the source data is sensitive or confidential. 
To address 
these concerns, source-free domain adaptation (SFDA) \cite{liang2020we, yang2021generalized} was proposed, where domain adaptation to the target distribution takes place with a source-pretrained model using only the unlabelled target data. 

While the vast majority of existing UDA literature deals with closed-set domain adaptation, where the target domain and source domain share the same classes, a more realistic scenario is open-set domain adaptation (OSDA) \cite{saito2018open, liu2019separate} where the target domain contains samples belonging to novel classes that are absent in the source domain. In the OSDA setting, closed-set UDA solutions would enforce alignment of the source and target feature spaces under the unknown category mismatch, leading to negative transfer \cite{fang2020open} and deteriorating performance. 
The majority of the existing OSDA methods \cite{saito2018open, liu2019separate} utilize domain adversarial learning techniques to align the source domain with only the known classes in the target domain, leaving out the target-unknown classes. 
Such methods fail to properly learn the features for the unknown classes, and hence no clear decision boundary between the known classes and the unknown class in the target domain is realized. 
Some universal domain adaptation methods, i.e. UDA methods designed to work in both closed and open-set settings \cite{saito2020universal, li2021domain}, have attempted to conduct self-supervised learning (SSL) to discover latent target domain features without explicit distribution matching. 
However, such methods fail under large domain gaps. 
More recently, \cite{jang2022unknown} proposed a three-way domain adversarial feature space alignment between the source domain and the known and the unknown target subdomains, thus segregating the known and unknown classes in the target domain.

In this work, we introduce \textbf{U}nknown \textbf{S}ample \textbf{D}iscovery (\textbf{USD}) as a source-free OSDA (SF-OSDA) method that utilizes an ensemble-based pseudolabeling strategy for the target data, and generates known and unknown target subsets based on Jensen-Shannon distance (JSD) between the pseudolabels and the predictions from a teacher model. 
USD uses 2-component Gaussian Mixture Model (GMM) to model the target domain JSD, where the distribution with the lower mean JSD is considered to be of the known class samples and that with the greater mean JSD is taken as that consisting of unknown class samples.
The known-unknown target subsets are used to adapt the student model.
The student model is updated with gradient descent, while the teacher model is updated by exponential moving averages (EMA) of the teacher and student models. 
The teacher-student framework in USD helps to mitigate error accumulation induced from any possibly faulty known-unknown sample separation.

USD introduces an unknown class output node in the target model. 
The adapted target model infers new target samples
in one one of the known classes or the unknown class, without operating on the entire target dataset first to identify known and unknown samples. 
The main contributions of this work are as follows.

\begin{itemize}
\item We introduce USD as an SF-OSDA model that co-trains a dual-branch teacher-student framework to split the target domain into known and unknown class subsets.
\item USD proposes the Jensen-Shannon distance between the target pseudolabels and teacher model predictions as an effective criterion for separating target samples in known and unknown classes.
\item Co-training in USD, aided by weak-strong consistency between the teacher and student outputs, significantly mitigates error accumulation resulting from imperfect known-unknown separation, and sustains the adaptation performance.
\item USD generates reliable pseudolabels from the student model outputs on an ensemble of weak and strong target data augmentations.
\item USD utilizes curriculum adaptation to progressively learn the known class feature space first, and the unknown class feature space later, thus enabling robust alignment of the entire target space with the source domain. 
\item Extensive experiments on 3 popular UDA benchmarks demonstrate the superiority of USD over existing SF-OSDA methods.
\end{itemize}

\section{Related Works}
\label{sec:related_works}

\subsection{Unsupervised domain adaptation}
Domain gap originates from the distribution shift between the source domain where a deep network model is trained, and the target domain where the model is deployed \cite{torralba2011unbiased}. 
This domain gap may be reduced by minimizing the maximum mean discrepancy (MMD) \cite{long2015learning, tzeng2014deep}, or the central moment discrepancy (CMD) \cite{zellinger2017central} between the distributions in the source and target domains. 
Deep CORAL \cite{sun2016deep} mitigated domain shift by matching second-order distribution statistics.
\cite{ganin2016domain} introduced the Gradient Reversal Layer (GRL) and made use of a domain discriminator to adversarially align the source and target distributions in a common feature space using a common feature encoder.
The Adversarial Discriminative Domain Adaptation (ADDA) \cite{tzeng2017adversarial} method decoupled the feature extraction process by learning two separate feature encoders for the two domains and aligned them adversarially to perform classification with a common classifier. 

Generative adversarial networks (GANs) have been utilized to produce images in an intermediate domain between the source and target to facilitate easier and smoother adaptation \cite{hoffman2018cycada}.
Domain-wise global adversarial alignment in the absence of target annotations may lead to loss of class discrimination in the target embeddings. 
To align  the domain-wise and class-wise distributions across the source and target data while maintaining target class feature discrimination, \cite{li2019joint} simultaneously solved two complementary domain-specific and class-specific minimax objectives.
The non-adversarial alignment approach in \cite{pan2019transferrable} imposed a consistency constraint between the labeled source prototypes and the pseudo-labeled target prototypes in the feature space. 

\subsection{Source free domain adaptation}
UDA methods that adversarially align the embedding space \cite{ganin2016domain, tzeng2017adversarial, hoffman2018cycada} or minimize the source-target domain divergence \cite{long2015learning, tzeng2014deep, zellinger2017central} require access to both the source and target data during adaptation, rendering them unusable in situations where the source data is private or restricted. 
A semi-supervised UDA method involving a few source representatives or prototypes instead of the full source data was proposed in \cite{chidlovskii2016domain}.
Distant supervision for SFDA \cite{liang2019distant} iteratively assigned pseudo-labels to the target data and used them to learn a domain invariant feature space and obtain the target class centroids.
Liang \textit{et al.} \cite{liang2020we} introduced SHOT which adapts the source-pretrained feature encoder to the target domain via self-training with information maximization \cite{krause2010discriminative, shi2012information} and self-supervised clustering for pseudolabeling, while transferring the source hypothesis (classifier model) to the target.
To further refine the pseudolabels, \cite{yang2021generalized} proposed to enforce neighborhood consistency regularization among the target samples. 
To generate compact target clusters, \cite{yang2022attracting} considered minimizing the distance among K-nearest neighbors for each target sample and dispersing the rest by retrieving target features stored in a memory bank.

\subsection{Open set domain adaptation}
In addition to aligning the source and target subspaces, a critical step in OSDA is detecting target samples from novel or unknown categories that are absent in the source domain. 
\cite{jain2014multi} applied a simple class-wise confidence threshold to reject those samples with lower confidence as unknown. 
\cite{saito2018open} adversarially aligned the source domain and known target subdomain, where the unknown target samples were identified based on a preset threshold. 
Alignment for only the known classes however results in subpar performance in identifying the unknown samples. 
The adversarial alignment objective was modified in \cite{liu2019separate} with an instance weighting procedure, where higher weights were given to known target samples and lower weight to unknown samples. 
This somewhat smoothened the known-unknown distinction, but lower weights produced less contributions in the objective loss from the unknown samples, leading to suboptimal performance.
A 3-way domain adversarial alignment between source, known target, and unknown target in the feature space was proposed in \cite{jang2022unknown} such that the source and known target are aligned while the target-unknown gets segregated. 
\cite{liang2020we} and \cite{yang2022attracting} are SF-UDA methods that also conduct SF-OSDA by separating the known and unknown samples based on clustering the sample entropies into two clusters, and taking the cluster with lower mean entropy as the known subset. 

\section{Method}
\label{sec:method}

For unsupervised OSDA, we have $n_s$ labeled samples ${\{x_s^i, y_s^i\}}_{i=1}^{n_s} \in {\mathcal{X}_s, \mathcal{Y}_s}$ belonging to the source domain $\mathcal{D}_s$, and $n_t$ unlabeled samples ${\{x_t^i\}}_{i=1}^{n_t} \in \mathcal{X}_t$ belonging to the target domain $\mathcal{D}_t$. 
The task of SF-OSDA is to take the source model $f_s (\theta_s): \mathcal{X}_s \rightarrow \mathcal{Y}_s$ with model parameters $\theta_s$ trained on the ${C_s}$-multiclass source data ${\{x_s^i, y_s^i\}}_{i=1}^{n_s} \in {\mathcal{X}_s, \mathcal{Y}_s}$, and adapt it to  $f_t (\theta_t): \mathcal{X}_t \rightarrow \mathcal{Y}_t$ with model parameters $\theta_t$ that can map the ${\{x_t^i\}}_{i=1}^{n_t} \in \mathcal{X}_t$ to the ${C_t}$ classes, where $C_t = C_s + 1$. 
The additional class in the target domain is a catch-all class for all samples in the target domain that do not belong to any of the classes in the source domain.

We follow \cite{liang2020we} for
Source model training follows \cite{liang2020we} to ensure fair comparison with other source-free UDA models. The source model is trained by minimizing the standard cross entropy loss with label smoothing \cite{muller2019does} as follows. 
\begin{equation}
    \mathcal{L}_{s}(f_s;\mathcal{X}_{s}, \mathcal{Y}_{s}) =  -\mathbb{E}_{x_s \in \mathcal{X}_s, y_s \in \mathcal{Y}_s} \sum_{k=1}^{C_s} q_k^{ls} \log(\sigma_k(f_s(x_s)))
\end{equation}
\noindent
where $\sigma_k(a) = \frac{exp(a_k)}{\sum_i exp(a_i)}$ is the k-th element in its softmax output of a $C_s$-dimensional vector $a$, and $q^{ls}$ is the one-hot encoded and smoothed $C_s$-dimensional vector for sample label $y_s^i$, such that $q_k^{ls} = (1-\alpha)q_k + \alpha/C_s$, where $q_k$ is $1$ for the correct class and $0$ for all other classes, and $\alpha$ is the smoothing factor set at $0.1$.

The source model $f_s$ consists of a feature extractor $g_s: \mathcal{X}_s \rightarrow \mathbb{R}^d$ and a $C_s$-class classifier $h_s: \mathbb{R}^d \rightarrow \mathbb{R}^{C_s}$, such that $f_s(x) = h_s(g_s(x))$. 
USD consists of a student target model $f_t^S (\theta_t^S)$ and a teacher target model $f_t^T (\theta_t^T)$.
The feature extractors $g_t^S$ and $g_t^T$, in the student and teacher networks respectively, are initialized with the source model feature extractor, i.e., $g_t^S=g_t^T=g_s$. 
To account for the novel class samples in the target domain, the source classifier $h_s$ is expanded in the student and teacher models to include an additional trainable output node for the unknown class. 
The known class nodes in the target classifiers $h_t^S$ and $h_t^T$, for the student and teacher respectively, are initialized with $h_s$, and remain frozen during adaptation.  
The unknown class nodes in $h_t^S$, $h_t^T$ and the feature extractors $g_t^S$, $g_t^T$ are adapted using only the unlabeled target samples.



\begin{figure}
\begin{minipage}[b]{1.0\linewidth}
  \centering
    \includegraphics[width=1\textwidth]{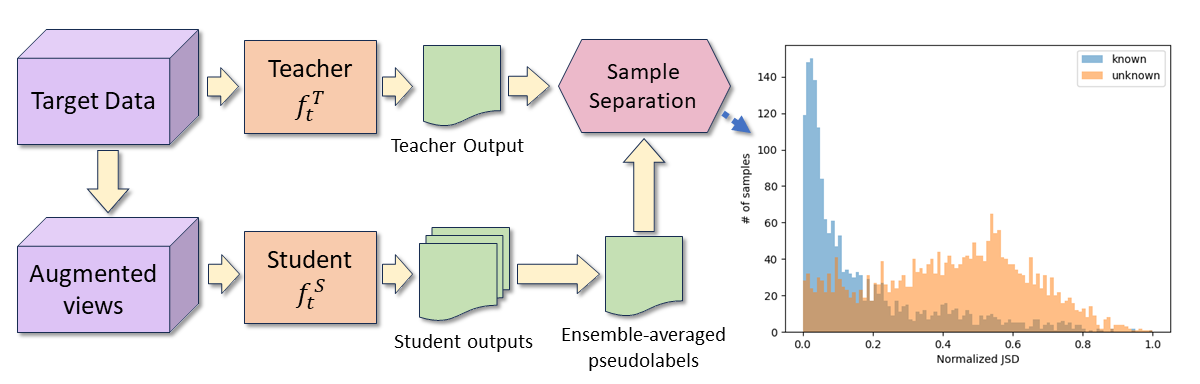}
\end{minipage}
    \caption{Pseudolabel generation for the target samples and known-unknown sample separation based on JSD}
    \label{fig:sample_separation}
\end{figure}

\begin{figure}
\begin{minipage}[b]{1.0\linewidth}
  \centering
    \includegraphics[width=1\textwidth]{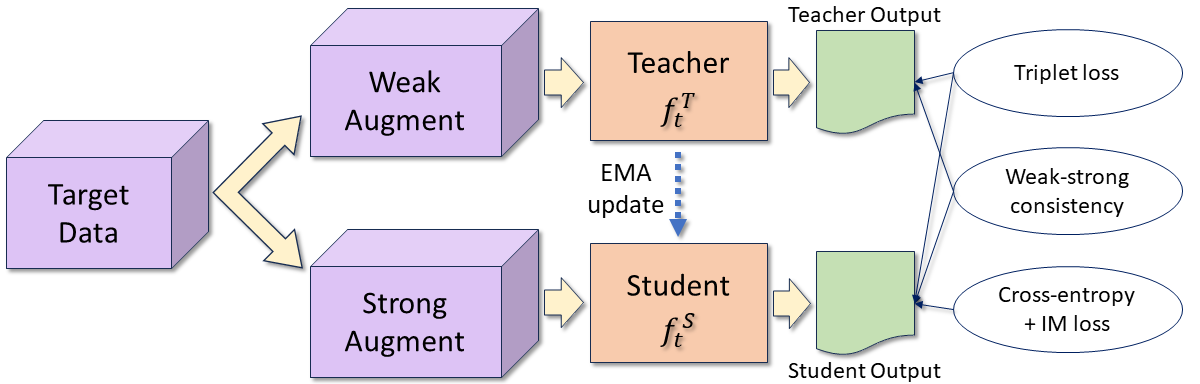}
\end{minipage}
    \caption{Adaptation process for USD using co-training. The student model receives pseudolabels for the target samples (see Figure \ref{fig:sample_separation})  and is optimized using a combination of triplet, weak-strong consistency, information maximization (IM) and cross-entropy losses.
    The teacher model is updated via exponential moving averages (EMA) at the end of each epoch.}
    \label{fig:method}
\end{figure}

\subsection{Known-unknown sample separation}

The first step for target adaptation is to reliably separate the known class samples and the novel class samples in the target data. 
This step is visually depicted in Figure \ref{fig:sample_separation}. 
In order to generate pseudolabels $\hat{y}_t$, the target data undergoes $M=6$ number of weak and strong augmentations (1 weak and 5 strong) based on \textit{AutoAugment} \cite{cubuk2019autoaugment} policy for ImageNet. 
The softmax output over $C_s$ classes for each augmented view $x_{t}^{iM}$ is taken from the student model $f_t^S$, and then averaged over the augmentations, as follows. 
\begin{equation}
    \hat{y}_t^i = \arg \max \frac{1}{M} \sum_1^{M} f_t^S(x_{t}^{iM})
    \label{eq:pseudolabel1}
\end{equation}
The index corresponding to the maximum averaged softmax output is taken as the hard pseudolabel $\hat{y}_t^i$ for each target sample $x_t^i$. 
These pseudolabels are however only over the $C_s$ known classes, and therefore the samples need to be split into known class subset $\mathcal{X}_t^K$ and unknown class subset $\mathcal{X}_t^U$. 
Existing SF-OSDA methods \cite{liang2020we, yang2022attracting} identify unknown class samples by utilizing the output entropy of the target data. 
Entropies for all samples are calculated at the beginning of each epoch and then normalized in the range of $[0,1]$ by dividing the each sample entropy by $\log C_s$. 
The normalized entropies are then clustered by 2-class k-means clustering. 
The cluster with the higher mean entropy or uncertainty is considered to be the one containing unknown samples, while the other cluster with lower mean entropy is taken as containing known class samples. 

Sample separation is  a critical component for noisy label learning (NLL) algorithms where clean and noisy samples are separated for robust supervised training of a model. 
Traditionally, NLL calculates the cross-entropy loss on the whole dataset and then uses low cross-entropy loss as the criterion to identify clean samples \cite{arazo2019unsupervised, dividemix, Li_2022_CVPR}. 
In USD, we conduct known-unknown sample separation for SF-OSDA based on JSD between the network outputs and their corresponding pseudolabels, which is calculated as follows.
\begin{equation}
    JSD(\hat{y}_t^i, p_t^i) = \\
    \frac{1}{2}KL \left ( \hat{y}_t^i, \frac{\hat{y}_t^i + p_t^i}{2} \right ) + \frac{1}{2}KL \left ( p_t^i, \frac{p_t^i + \hat{y}_t^i}{2} \right )
    \label{eq:1}
\end{equation}
where, $KL(a,b)$ is the Kullback-Leibler divergence between $a$ and $b$, and $p_t^i=\sigma(f_t^T(x_t^i))$ is the output softmax probability for target sample $x_t^i$ from the target teacher model $f_t^T$.

We consider the unknown class samples in the target domain as noisy samples when predictions are made over only the known $C_s$ classes. 
In comparison to entropy or cross-entropy loss, JSD is symmetric by design and ranges between 0 and 1.
As shown in Figure \ref{fig:sample_separation}, when plotted against the number of samples, JSD produces a bimodal histogram. 
We model the JSD distribution with 2-component Multivariate Gaussian Mixture Model (GMM) with equal priors, resulting in probabilities for each target sample to belong to either of the two modes.
We consider the samples belonging to the distribution with the lower-mean Gaussian as samples from one of the known classes, and consider those samples on the higher-mean Gaussian as coming from the unknown target class.

Practically, we take the probability $w_t^i$ of belonging to the lower-mean GMM distribution for each target sample $x_t^i$, and set a lower-bound/threshold $\delta_t$ to select the known sample subset $\mathcal{X}_t^K$. 
The remaining target samples are included in the unknown subset $\mathcal{X}_t^U$. 
The pseudolabels $\hat{y}_t^i$ are updated accordingly, where the known subset retain their earlier assigned pseudolabel from among the $C_s$ classes, and the unknown subset of target samples get the new unknown class pseudolabel $|C_t|$. 
It has to be noted that during adaptation, the teacher network conducts the known-unknown sample separation at the beginning of each epoch, and the student network is adapted over the $C_t$ classes with the target data. 

\subsection{Teacher-student co-training and regularization}

USD simultaneously adapts the student and teacher target models, such that the student model parameters $\theta_t^S$ are updated based on the minibatch gradient descent, and the teacher network parameters $\theta_t^T$ are updated as temporally ensembled version  of the student network \cite{tarvainen2017mean} at the end of each epoch as follows. 

\begin{equation}
    \theta_{t_N}^T = m \theta_{t_{N-1}}^T + (1-m) \theta_{t_N}^S
    \label{eq:teacher_update}
\end{equation}
\noindent
where, $m$ is the momentum parameter for weight ensembling, and $N = 2, 3, .., E$ is the epoch number.
Such co-training and cross-network sample splitting by the teacher for the student work to lessen error accumulation from imperfect known-unknown sample separation and stabilizes the adaptation process.
USD further maintains weak-strong temporal consistency between the teacher network outputs and the student network outputs by minimizing the following consistency loss.
\begin{equation}
    \mathcal{L}_{t}^{con}(f_t^S, f_t^T;\mathcal{X}_{t}) =  KL \left ( p_t^{iS}, p_t^{iT} \right ) =  \sum_{k=1}^{C_t} p_t^{iT} \log  \left ( \frac{p_t^{iT}}{p_t^{iS}} \right )
    \label{eq:consistency_loss}
\end{equation}
\noindent
where, $p_t^{iS} = \sigma(f_t^S(x_t^{iS}))$ is the softmax output from the student on an strongly augmented target sample $x_t^{iS}$, and $p_t^{iT} = \sigma(f_t^T(x_t^{iW}))$ is the softmax output from the teacher on the weakly augmented version $x_t^{iW}$ of the same target instance. The strong and weak augmentations are done following the \textit{AutoAugment} \cite{cubuk2019autoaugment} ImageNet policy. 

USD also utilizes a triplet loss \cite{schroff2015facenet} to effectively learn the decision boundary between known and unknown classes.
The output $z_T^{ia} = [f_t^T(x_t^{iW})]^a$ of the teacher model on an weakly augmented known class sample is taken as the anchor, and the corresponding output $z_S^{i+} = [f_t^S(x_t^{iS})]^+$ on the strongly augmented version of the same sample from the student model is taken as the positive instance. 
The negative instance is the student model output $z_S^{i-} = [f_t^S(x_t^{iS})]^-$ on a randomly chosen unknown class sample. 
Cosine distance is taken as the distance metric, and is calculated as follows.
\begin{equation}
    \mathbf{D}(z_1, z_2) = 1 - \frac{z_1 .z_2}{{||z_1||}_2 {||z_2||}_2}
\end{equation}
where $z_1$ and $z_2$ are any two network outputs. Triplet loss is in turn calculated as follows. 
\begin{equation}
    \mathcal{L}_{t}^{trip}(f_t^S, f_t^T;\mathcal{X}_{t}) = \max ( \mathbf{D}(z_T^{ia}, z_S^{i+}) - \mathbf{D}(z_T^{ia}, z_S^{i-}), 0)
    \label{eq:triplet_loss}
\end{equation}

In addition, the student network is trained with the instance-weighted  standard cross-entropy loss with label smoothing \cite{muller2019does}, as follows.
\begin{equation}
    \mathcal{L}_{t}^{ce}(f_t^S;\mathcal{X}_t) = -\mathbb{E}_{x_t^i \in \mathcal{X}_{t}} \omega^i \sum_{k=1}^{C_t} \hat{y}_{t_k}^i log(\sigma_k(f_t^S(x_t^{iS})))
    \label{eq:L_t_ce} 
\end{equation}
 
The instance weights $\omega^i$ are the probability $w_t^i$ for known target samples $x_i^t \in \mathcal{X}_t^K$ of belonging to the lower-mean JSD distribution, and $(1 - w_t^i)$ for unknown target samples $x_i^t \in \mathcal{X}_t^U$ of belonging to the higher-mean JSD distribution, during the known-unknown sample separation.
In order to promote adaptation to the known samples first and to progressively learn the unknown class feature space, USD utilizes cross-entropy loss under curriculum guidance, dictated by the curriculum factor $\gamma_r$ as follows.
\begin{multline}
    \mathcal{L}_{t}^{ce}(f_t^S;\mathcal{X}_t^K,\mathcal{X}_t^U) = \\ \gamma_n \mathcal{L}_{tK}^{ce}(f_t^S;\mathcal{X}_t^K) + (1-\gamma_r)\mathcal{L}_{tU}^{ce}(f_t^S;\mathcal{X}_t^U)
    \label{eq:L_t_ce_cur} 
\end{multline}
where $\gamma_r = \max(0.5, \gamma_{r-1} (1-\beta \epsilon^{-\mathcal{L}_{tK_r}^{ce}/\mathcal{L}_{tK_{r-1}}^{ce}}))$ such that, $\beta$ is a hyperparameter and $r$ is the current iteration number. 
The ratio $\mathcal{L}_{tK_r}^{ce}/\mathcal{L}_{tK_{r-1}}^{ce}$ dictates the degree by which the curriculum factor decreases from the earlier $(r-1)$-th iteration to the current $r$-th iteration. 
When loss $\mathcal{L}_{tK}^{ce}$ on the known sample subset increases, $\gamma$ marginally decreases to accommodate further adaptation on the known samples in the subsequent iterations. 
But if $\mathcal{L}_{tK}^{ce}$ decreases by a large margin, $\gamma$ decreases accordingly to progressively adapt to the unknown samples in the following iterations.
Curriculum guidance balances the adaptation of the target model to the known and unknown subsets.

To encourage individually precise and globally diverse predictions, USD further minimizes the information maximization (IM) \cite{liang2020we} loss as formulated in \cite{taufique2023continual, taufique2022unsupervised}.
\begin{multline}
    \mathcal{L}_t^{ent}(f_t^S;\mathcal{X}_{t}^K) \\ = -\mathbb{E}_{x_t^i \in \mathcal{X}_{t}^K} \sum_{k=1}^{C_t} \sigma_k(f_t^S(x_t^{iS}))log(\sigma_k(f_t^S(x_t^{iS})))
    \label{eq:ent}
\end{multline}
\begin{equation}
    \mathcal{L}_t^{eqdiv}(f_t^S;\mathcal{X}_{t}^K) = \sum_{k=1}^{C_t} p_t^{iS} log \left ( \frac{p_t^{iS}}{\overline{{p_t^{iS}}}} \right )
    \label{eq:eqdiv}
\end{equation}
\begin{equation}
    \mathcal{L}_t^{IM}(f_t^S;\mathcal{X}_{t}^K) = \mathcal{L}_t^{ent}(f_t^S;\mathcal{X}_{t}^K) + \mathcal{L}_t^{eqdiv}(f_t^S;\mathcal{X}_{t}^K)
    \label{eq:IM}
\end{equation}
where $\overline{{p_t^{iS}}} = \mathbb{E}_{x_t^i \in \mathcal{X}_{t}^K} [ \sigma(f_t^S(x_t^{iS})) ]$ is the mean softmax output vector over known target samples in a minibatch. The overall objective function is therefore,
\begin{equation}
    \mathcal{L}_t^{tot} = \mathcal{L}_{t}^{ce} + \mathcal{L}_t^{IM} + \zeta_1 \mathcal{L}_{t}^{trip} + \zeta_2 \mathcal{L}_{t}^{con} 
    \label{eq:tot}
\end{equation}
where $\zeta_1$ and $\zeta_2$ are two hyperparameters.

A brief demonstration of the USD domain adaptation pipeline is presented in Algorithm \ref{alg:algo}. 

\begin{algorithm}[!ht]
\caption{Pseudocode for USD}\label{alg:algo}
\DontPrintSemicolon
  \KwInput{Source trained model $f_s$ and $n_t$ unlabled target data samples $x_t^i \in \mathcal{X}_{t}$}
  \KwOutput{Target adapted student model $f_t^S$}
  \KwInit{Teacher target model $f_{t}^T$ and student target model $f_{t}^S$, are both initialized with parameters $\theta_s$ from $f_s$}
  \For{epoch = $1$ to $E$}
    {
    Conduct $M=6$ weak-strong augmentations and assign ensemble averaged pseudolabels $\hat{y}_t^i$ using eq.~(\ref{eq:pseudolabel1}) \;
    Conduct known ($\mathcal{X}_{t}^K$) - unknown ($\mathcal{X}_{t}^U$) target sample separation using JSD between $\hat{y}_t^i$ and teacher  softmax output $p_t^i = \sigma(f_t^T(x_t^i))$ \; 
    \For{$i$ = $1$ to $n_t$}
        {
        Optimize, for each minibatch, student model $f_t^S$ with loss $\mathcal{L}_{tot}$ using eq.~(\ref{eq:tot}) 
        and get new student model parameters $\theta_t^S$
        }
    Update teacher model $f_t^T$ using new student model weights $\theta_t^S$ and current teacher model weights $\theta_t^T$ using eq.~(\ref{eq:teacher_update})   
    }
\end{algorithm}
\section{Experimental Setup}
\label{sec:experimental_setup}

\subsection{Datasets}
We evaluate USD on three popular domain adaptation benchmarks: Office-31 \cite{saenko2010adapting}, Office-Home \cite{venkateswara2017deep}, and VisDA-C \cite{peng2018visda}.
\textbf{Office-31} is a small-scale DA dataset with 3 distinct domains, Amazon (A), Webcam (W), and DSLR (D), consisting of images belonging to 31 classes of objects found generally in an office environment.
Divided into 4 distinct domains Art (A), Clipart (C), Product (P), and Real-World (R), \textbf{Office-Home} is a medium-sized DA dataset which has images of 65 classes of objects found in contemporary office and home settings.
The large-scale \textbf{VisDA-C} dataset contains images of 12 classes of itemss over 2 domains: Synthetic (S) and Real (R). Its source domain is composed of 152K synthetically rendered 3D images. The target domain consists of 55K real images taken from MS COCO dataset \cite{lin2014microsoft}.
For OSDA, we follow the shared and target-private dataset splits done in \cite{saito2018open}. 

\subsection{Implementation details}
For source training, we follow the protocol from \cite{liang2020we, yang2022attracting} for fair comparison against existing SF-OSDA methods. 
The basic structure of the teacher and student models also follow that of \cite{liang2020we, yang2022attracting}, that is, the feature extractor is a ResNet-50 \cite{he2016deep}, followed by a fully-connected (FC) bottleneck layer, a batch normalization layer \cite{ioffe2015batch}, an FC classifier layer, and a weight normalization layer \cite{salimans2016weight}, respectively. 
The student target model is trained with an SGD optimizer with momentum of 0.9 and weight decay of $10^ {-3}$. 
Due to the difference in the number of samples in each dataset, USD is adapted for 40 epochs on Office and for 20 epochs on Office-Home at $\beta=0.01$, and for 5 epochs on VisDA-C at $\beta=0.001$, at minibatch size of 64 samples in all cases. 
The threshold $\delta_t$ for known-unknown sample separation is set at 0.8, and the momentum parameter $m$ for temporal ensembling is set according to schedule in \cite{Xu_2021_CVPR} with a maximum at 0.9995. 
Further, $\zeta_1=0.01$ and $\zeta_2$ is gradually increased to $0.5$ following \cite{laine2017temporal}. 
All experiments were done on a NVIDIA A100 GPU.

\subsection{Evaluation metrics}
The mean-per-class accuracy \textbf{OS} over all known classes and the unified unknown class for all the target data may be considered as a metric for evaluating OSDA. 
However, such a metric is dominated by the accuracy on the known classes, as all the unknown samples are lumped into 1 unknown class \cite{bucci2020effectiveness}. 
A better metric is therefore to calculate the mean-per-class accuracy \textbf{OS*} over only the known classes, and the accuracy \textbf{UNK} for the unknown class, and then take the harmonic mean \textbf{HOS} of the two for fair evaluation over the known and the unknown classes. 
Mathematically, the metrics are formulated as follows.
\begin{equation}
    OS^{*} = \frac{1}{|C_s|} \sum_{i=1}^{|C_s|} \frac{|x_t:x_t \in \mathcal{D}_t^i \cap \Tilde{y}_t^i=i|}{|x_t:x_t \in \mathcal{D}_t^i|}
\end{equation}
\begin{equation}
    UNK = \frac{|x_t:x_t \in \mathcal{D}_t^{|C_t|} \cap \Tilde{y}_t^i=|C_t||}{|x_t:x_t \in \mathcal{D}_t^{|C_t|}|}
\end{equation}
\begin{equation}
    HOS = \frac{2 \times OS^* \times UNK}{OS^* + UNK}
\end{equation}
Here, $\Tilde{y}_t^i = \arg \max (\sigma(f_t^S(x_t^i)))$ is the prediction from the student model $f_t^S$ and $\mathcal{D}_t^i$ is the target domain data belonging to class $i$. In this work, we report OS*, UNK, and HOS for the evaluated adaptation tasks.
\section{Results}
\label{sec:results}

\begin{table*}[!htbp]
\begin{center}
\resizebox{\textwidth}{!}{
\begin{tabular}{L|c|ccc|ccc|ccc|ccc|ccc|ccc|cca}
\toprule
    \multirow{2}{*}{Method} & \multirow{2}{*}{SF} & \multicolumn{3}{c}{A $\rightarrow$ D} & \multicolumn{3}{c}{A $\rightarrow$ W} & \multicolumn{3}{c}{D $\rightarrow$ A} & \multicolumn{3}{c}{D $\rightarrow$ W} & \multicolumn{3}{c}{W $\rightarrow$ A} & \multicolumn{3}{c}{W $\rightarrow$ D} & 
    \multicolumn{3}{c}{Avg.} \\ 
    &  & OS* & UNK & HOS & OS* & UNK & HOS & OS* & UNK & HOS & OS* & UNK & HOS & OS* & UNK & HOS & OS* & UNK & HOS & OS* & UNK & HOS \\ 
    \midrule
    DANN\cite{ganin2016domain} & \tikzxmark & 90.8 & 59.2 & 71.5 & 87.4 & 55.7 & 68.1 & 72.9 & 74.5 & 73.7 & 99.3 & 77.0 & 86.7 & 72.1 & 73.1 & 72.6 & 100.0 & 70.2 & 82.5 & 87.1 & 68.3 & 75.9 \\ 
    CDAN\cite{long2018conditional} & \tikzxmark & 92.2 & 52.4 & 66.8 & 90.3 & 50.7 & 64.9 & 74.9 & 70.6 & 72.7 & 99.6 & 73.2 & 84.3 & 72.8 & 69.3 & 71.0 & 100.0 & 67.3 & 80.5 & 88.3 & 63.9 & 73.4 \\
    STA\cite{liu2019separate} & \tikzxmark & 91.0 & 63.9 & 75.0 & 86.7 & 67.6 & 75.9 & 83.1 & 65.9 & 73.2 & 94.1 & 55.5 & 69.8 & 66.2 & 68.0 & 66.1 & 84.9 & 67.8 & 75.2 & 84.3 & 64.8 & 72.5 \\
    OSBP\cite{saito2018open} & \tikzxmark & 90.5 & 75.5 & 82.4 & 86.8 & 79.2 & 82.7 & 76.1 & 72.3 & 75.1 & 97.7 & 96.7 & 97.2 & 73.0 & 74.4 & 73.7 & 99.1 & 84.2 & 91.1 & 87.2 & 80.4 & 83.7 \\
    PGL\cite{luo2020progressive} & \tikzxmark & 82.1 & 65.4 & 72.8 & 82.7 & 67.9 & 74.6 & 80.6 & 61.2 & 69.5 & 87.5 & 68.1 & 76.5 & 80.8 & 61.8 & 70.1 & 82.8 & 64.0 & 72.2 & 82.7 & 64.7 & 72.6 \\
    OSLPP\cite{wang2021progressively} & \tikzxmark & 92.6 & 90.4 & 91.5 & 89.5 & 88.4 & 89.0 & 82.1 & 76.6 & 79.3 & 96.9 & 88.0 & 92.3 & 78.9 & 78.5 & 78.7 & 95.8 & 91.5 & 93.6 & 89.3 & 85.6 & 87.4 \\
    UADAL\cite{jang2022unknown} & \tikzxmark & 85.1 & 87.0 & 86.0 & 84.3 & 94.5 & 89.1 & 73.3 & 87.3 & 79.7 & 99.3 & 96.3 & 97.8 & 67.4 & 88.4 & 76.5 & 99.5 & 99.4 & 99.5 & 84.8 & 92.1 & 88.1 \\
    \midrule
    SHOT*\cite{liang2020we} & \tikzcmark & 94.0 & 46.3 & 62.0 & 95.6 & 42.3 & 58.7 & 83.3 & 39.1 & 53.3 & 100.0 & 75.7 & 86.1 & 82.7 & 46.6 & 59.6 & 100.0 & 69.7 & 82.1 & 92.6 & 53.3 & 67.0 \\
    AaD*\cite{yang2022attracting} & \tikzcmark & 73.0 & 84.6 & 78.3 & 63.5 & 89.5 & 74.3 & 63.6 & 88.9 & 74.2 & 78.0 & 98.5 & 87.0 & 61.9 & 88.9 & 73.0 & 94.6 & 96.8 & 95.7 & 72.4 & 91.2 & 80.4 \\
    \rowcolor{maroon!10}
    USD (Ours) & \tikzcmark & 90.7 & 73.4 & 81.2 & 82.8 & 72.7 & 77.9 & 65.7 & 84.4 & 73.9 & 97.9 & 96.6 & 97.3 & 64.6 & 86.7 & 74.0 & 98.0 & 92.6 & 95.2 & 83.3 & 84.4 & 83.3 \\
\bottomrule
\end{tabular}}
\end{center}
\caption{Evaluation of USD on Office-31 dataset. * are results computed for the methods using publicly released code.}
\label{res:result_office}
\end{table*}

\begin{table*}[!htbp]
\resizebox{1\textwidth}{!}{
\begin{tabular}{L|c|ccc|ccc|ccc|ccc|ccc|ccc|ccc}
\toprule
    \multirow{3}{*}{Method} & \multirow{3}{*}{SF} & \multicolumn{21}{c}{Office-Home} \\
    \cline{3-23}
    & & \multicolumn{3}{c}{A $\rightarrow$ C} & \multicolumn{3}{c}{A $\rightarrow$ P} & \multicolumn{3}{c}{A $\rightarrow$ R} & \multicolumn{3}{c}{C $\rightarrow$ A} & \multicolumn{3}{c}{C $\rightarrow$ P} & \multicolumn{3}{c}{C $\rightarrow$ R} & \multicolumn{3}{c}{P $\rightarrow$ A} \\
    &  & OS* & UNK & HOS & OS* & UNK & HOS & OS* & UNK & HOS & OS* & UNK & HOS & OS* & UNK & HOS & OS* & UNK & HOS & OS* & UNK & HOS \\
    \midrule 
    DANN\cite{ganin2016domain} & \tikzxmark & 37.1 & 82.7 & 51.2 & 60.0 & 71.3 & 65.2 & 75.1 & 67.3 & 71.0 & 43.8 & 84.3 & 57.6 & 50.1 & 77.6 & 60.9 & 61.1 & 73.5 & 66.7 & 42.4 & 83.9 & 56.3 \\
    CDAN\cite{long2018conditional} & \tikzxmark & 39.7 & 78.9 & 52.9 & 61.7 & 68.8 & 65.1 & 75.2 & 66.7 & 70.7 & 44.9 & 82.8 & 58.2 & 51.6 & 76.8 & 61.7 & 61.5 & 73.7 & 67.1 & 45.8 & 81.2 & 58.6 \\ 
    STA\cite{liu2019separate} & \tikzxmark & 46.0 & 72.3 & 55.8 & 68.0 & 48.4 & 54.0 & 78.6 & 60.4 & 68.3 & 51.4 & 65.0 & 57.4 & 61.8 & 59.1 & 60.4 & 67.0 & 66.7 & 66.8 & 54.2 & 72.4 & 61.9 \\
    OSBP\cite{saito2018open} & \tikzxmark & 50.2 & 61.1 & 55.1 & 71.8 & 59.8 & 65.2 & 79.3 & 67.5 & 72.9 & 59.4 & 70.3 & 64.3 & 67.0 & 62.7 & 64.7 & 72.0 & 69.2 & 70.6 & 59.1 & 68.1 & 63.2  \\
    PGL\cite{luo2020progressive} & \tikzxmark & 63.3 & 19.1 & 29.3 & 78.9 & 32.1 & 45.6 & 87.7 & 40.9 & 55.8 & 85.9 & 5.3 & 10.0 & 73.9 & 24.5 & 36.8 & 70.2 & 33.8 & 45.6 & 73.7 & 34.7 & 47.2 \\
    OSLPP\cite{wang2021progressively} & \tikzxmark & 55.9 & 67.1 & 61.0 & 72.5 & 73.1 & 72.8 & 80.1 & 69.4 & 74.3 & 49.6 & 79.0 & 60.9 & 61.6 & 73.3 & 66.9 & 67.2 & 73.9 & 70.4 & 54.6 & 76.2 & 63.6 \\
    UADAL\cite{jang2022unknown} & \tikzxmark & 54.9 & 74.7 & 63.2 & 69.1 & 72.5 & 70.8 & 81.3 & 73.7 & 77.4 & 53.5 & 80.5 & 64.2 & 62.1 & 78.8 & 69.5 & 69.1 & 78.3 & 73.4 & 50.5 & 83.7 & 63.0 \\
    \midrule 
    SHOT\cite{liang2020we} & \tikzcmark & 67.0 & 28.0 & 39.5 & 81.8 & 26.3 & 39.8 & 87.5 & 32.1 & 47.0 & 66.8 & 46.2 & 54.6 & 77.5 & 27.2 & 40.2 & 80.0 & 25.9 & 39.1 & 66.3 & 51.1 & 57.7 \\
    AaD\cite{yang2022attracting} & \tikzcmark & 50.7 & 66.4 & 57.6 & 64.6 & 69.4 & 66.9 & 73.1 & 66.9 & 69.9 & 48.2 & 81.1 & 60.5 & 59.5 & 63.5 & 61.4 & 67.4 & 68.3 & 67.8 & 47.3 & 82.4 & 60.1 \\
    \rowcolor{maroon!10}
    USD (Ours) & \tikzcmark & 53.3	& 71.5 & 61.1 & 65.7 & 74.9 & 70 & 73.3 & 79.5 & 76.3 & 52.2 & 70.8 & 60.1 & 62.4 & 68.4 & 65.2 & 69.3 & 68.6 & 68.9 & 54.3 & 73.8 & 62.6 \\
\bottomrule
\toprule
    \multirow{3}{*}{Method} & \multirow{3}{*}{SF} & \multicolumn{18}{c}{Office-Home} & \multicolumn{3}{c}{\multirow{2}{*}{VisDA-C}} \\
    \cline{3-20}
    & & \multicolumn{3}{c}{P $\rightarrow$ C} & \multicolumn{3}{c}{P $\rightarrow$ R} & \multicolumn{3}{c}{R $\rightarrow$ A} & \multicolumn{3}{c}{R $\rightarrow$ C} & \multicolumn{3}{c}{R $\rightarrow$ P} & \multicolumn{3}{c}{Avg.} \\ 
    &  & OS* & UNK & HOS & OS* & UNK & HOS & OS* & UNK & HOS & OS* & UNK & HOS & OS* & UNK & HOS & OS* & UNK & \cellcolor{lightmintbg}HOS & OS* & UNK & \cellcolor{lightmintbg}HOS \\
    \midrule 
    DANN\cite{ganin2016domain} & \tikzxmark & 30.1 & 86.3 & 44.6 & 67.7 & 72.0 & 69.8 & 56.8 & 77.1 & 65.4 & 37.1 & 80.9 & 50.9 & 69.6 & 67.2 & 68.4 & 52.6 & 77.1 & \cellcolor{lightmintbg}60.7 & 52.1 & - & \cellcolor{lightmintbg}-\\ 
    CDAN\cite{long2018conditional} & \tikzxmark & 33.1 & 82.4 & 47.2 & 69.8 & 69.7 & 69.7 & 59.8 & 73.6 & 66.0 & 40.3 & 75.8 & 52.7 & 70.9 & 64.6 & 67.6 & 54.5 & 74.6 & \cellcolor{lightmintbg}61.4 & - & - & \cellcolor{lightmintbg}- \\ 
    STA\cite{liu2019separate} & \tikzxmark & 44.2 & 67.1 & 53.2 & 76.2 & 64.3 & 69.5 & 67.5 & 66.7 & 67.1 & 49.9 & 61.1 & 54.5 & 77.1 & 55.4 & 64.5 & 61.8 & 63.3 & \cellcolor{lightmintbg}61.1 & 62.4 & 82.4 & \cellcolor{lightmintbg}71.0\\
    OSBP\cite{saito2018open} & \tikzxmark & 44.5 & 66.3 & 53.2 & 76.2 & 71,7 & 73.9 & 66.1 & 67.3 & 66.7 & 48.0 & 63.0 & 54.5 & 76.3 & 68.6 & 72.3 & 64.1 & 66.3 & \cellcolor{lightmintbg}64.7 & 50.9 & 81.7 & \cellcolor{lightmintbg}62.7\\
    PGL\cite{luo2020progressive} & \tikzxmark & 59.2 & 38.4 & 46.6 & 84.8 & 27.6 & 41.6 & 81.5 & 6.1 & 11.4 & 68.8 & 0.0 & 0.0 & 84.8 & 38.0 & 52.5 & 76.1 & 25.0 & \cellcolor{lightmintbg}35.2 & - & - & \cellcolor{lightmintbg}- \\
    OSLPP\cite{wang2021progressively} & \tikzxmark & 53.1 & 67.1 & 59.3 & 77.0 & 71.2 & 74.0 & 60.8 & 75.0 & 67.2 & 54.4 & 64.3 & 59.0 & 78.4 & 70.8 & 74.4 & 63.8 & 71.7 & \cellcolor{lightmintbg}67.0 & - & - & \cellcolor{lightmintbg}- \\
    UADAL\cite{jang2022unknown} & \tikzxmark & 43.4 & 81.5 & 56.6 & 71.6 & 83.1 & 76.9 & 66.7 & 78.6 & 72.1 & 51.1 & 74.5 & 60.6 & 77.4 & 76.2 & 76.8 & 62.6 & 78.0 & \cellcolor{lightmintbg}68.7 & - & - & \cellcolor{lightmintbg}- \\
    \midrule 
    SHOT\cite{liang2020we} & \tikzcmark & 59.3 & 31.0 & 40.8 & 85.8 & 31.6 & 46.2 & 73.5 & 50.6 & 59.9 & 65.3 & 28.9 & 40.1 & 84.4 & 28.2 & 42.3 & 74.6 & 33.9 & \cellcolor{lightmintbg}45.6 & 57.5* & 12.1* & \cellcolor{lightmintbg}20.1* \\
    AaD\cite{yang2022attracting} & \tikzcmark & 45.4 & 72.8 & 55.9 & 68.4 & 72.8 & 70.6 & 54.5 & 79.0 & 64.6 & 49.0 & 69.6 & 57.5 & 69.7 & 70.6 & 70.1 & 58.2 & 71.9 & \cellcolor{lightmintbg}63.6 & 32.0* & 62.9* & \cellcolor{lightmintbg}42.4* \\
    \rowcolor{maroon!10}
    USD (Ours) & \tikzcmark & 47.3 & 69.6 & 56.3 & 70 & 74.5 & 72.2 & 64.6 & 71.3 & 67.8 & 53.8 & 65.5 & 59.1 & 73.3 & 69.1 & 71.1 & 61.6 & 71.5 & 65.9 & 57.8 & 86.7 & 69.4 \\
\bottomrule
\end{tabular}
}
\caption{Evaluation of USD on Office-Home and VisDA-C datasets. * are results computed for the methods using publicly released code.}
\label{res:result_officehome}
\end{table*}



\subsection{Overall results}

We compare USD to a number of existing UDA methods: closed-set UDA methods (1) DANN \cite{ganin2016domain}, (2) CDAN \cite{long2018conditional}, open-set UDA methods (3) STA \cite{liu2019separate}, (4) OSBP \cite{saito2018open}, (5) PGL \cite{luo2020progressive}, (6) OSLPP \cite{wang2021progressively}, and (7) UADAL \cite{jang2022unknown}. 
These methods however are not source-free. 
We compare USD to open-set versions of SF-UDA methods SHOT \cite{liang2020we} and AaD \cite{yang2022attracting}. 
The open-set results for SHOT and AaD on Office-Home are provided in their respective publications. We generate results for Office-31 and VisDA-C using their publicly released code.

The results on Office-31 over all 6 domain pairs are presented in Table \ref{res:result_office}. 
USD outperforms SHOT and AaD by $\sim16\%$ and $\sim3\%$, respectively in terms of mean HOS. 
Distinguishing between known and unknown class samples is crucial in OSDA, and USD strikes the best balance among the other SF-OSDA methods. 
SHOT clearly adapts primarily to the known classes without good adaptation on the unknown samples. 
AaD overcompensates in identifying unknown samples at the expense of correctly adapting to the known classes. 
USD performs equally well over both known and unknown classes, leading to higher HOS. 
USD also outperforms non-source-free methods STA and PGL, while being comparable to OSBP.  

A comparative evaluation for USD against existing UDA methods on Office-Home is given in Table \ref{res:result_officehome}. 
USD outperforms SHOT and AaD by $\sim20\%$ and $\sim2\%$, respectively in terms of the average HOS over the 12 domain pairs. 
Similar to Office-31, SHOT adapts better to the known classes, but fails to competently identify unknown samples, while AaD performs worse on the known classes and better on the unknown samples. 
USD is more balanced across the known and unknown classes and also outperforms non-SF OSDA methods STA, OSBP and PGL.

Results on VisDA-C are given in the bottom right section in Table \ref{res:result_officehome}. SHOT severely suffers from negative transfer in the unknown class, while AaD fails to learn the target-known feature space. 
USD greatly outperforms SHOT and AaD, as well as the non-SF method OSBP, while being comparable to STA in terms of mean HOS.
 
\subsection{Ablation study}

\begin{table*}[thbp]
\begin{adjustbox}{max width=\textwidth}
\begin{tabular}{L|L|ccc|ccc|ccc|ccc|ccc|ccc|cca||cca}
\cline{1-26}
    \multirow{3}{*}{\makecell{Separation \\ criterion}} & \multirow{3}{*}{\makecell{Distribution \\ modeling}} & \multicolumn{21}{c}{Office} & \multicolumn{3}{c}{\multirow{2}{*}{VisDA-C}} \\
    \cline{3-23}
    & & \multicolumn{3}{c}{A $\rightarrow$ D} & \multicolumn{3}{c}{A $\rightarrow$ W} & \multicolumn{3}{c}{D $\rightarrow$ A} & \multicolumn{3}{c}{D $\rightarrow$ W} & \multicolumn{3}{c}{W $\rightarrow$ A} & \multicolumn{3}{c}{W $\rightarrow$ D} & 
    \multicolumn{3}{c}{Avg.}  \\ 
    &  & OS* & UNK & HOS & OS* & UNK & HOS & OS* & UNK & HOS & OS* & UNK & HOS & OS* & UNK & HOS & OS* & UNK & HOS & OS* & UNK & HOS & OS* & UNK & HOS\\ 
    \cline{1-26}
    \rowcolor{maroon!10}
    JSD & GMM & 89.4 &  70.2 &  78.6 & 82.7 & 73.0 & 77.6 & 66.4 & 85.2 & 74.6 & 97.5 & 97.0 & 97.2 & 68.3 & 85.4 & 75.9 & 98.0 & 93.6 & 95.8 & 83.7 & 84.1 & 83.3 & 57.8 & 86.7 & 69.4 \\
    Entropy & GMM & 88.9 & 70.2 & 78.4 & 83.3 & 74.5 & 78.6 & 65.3 & 90.5 & 75.9 & 97.9 & 93.3 & 95.5 & 60.2 & 88.5 & 71.7 & 98.0 & 93.1 & 95.5 & 82.3 & 85.0 & 82.6 & 57.1 & 85.4 & 68.4\\
    CE & GMM & 90.7 & 68.6 & 78.1 & 90.0 & 61.8 & 73.3 & 69.6 & 81.0 & 74.9 & 98.2 & 93.3 & 95.6 & 68.5 & 86.0 & 76.2 & 98.0 & 90.4 & 94.1 & 85.8 & 80.2 & 82.0 & 67.3 & 45.5 & 54.3\\
    \cline{1-26} 
    JSD & BMM & 91.4 & 53.7 & 67.7 & 93.6 & 53.2 & 67.8 & 77.7 & 72.3 & 74.9 & 100.0 & 82.4 & 90.3 & 77.1 & 72.5 & 74.8 & 100.0 & 71.3 & 83.2 & 90.0 & 67.6 & 76.5 & 67.6 & 58.3 & 62.6 \\
    Entropy & BMM & 90.2 & 60.1 & 72.1 & 87.2 & 78.3 & 82.5 & 66.3 & 88.5 & 75.8 & 89.5 & 92.1 & 90.8 & 60.8 & 87.1 & 71.6 & 100.0 & 88.8 & 94.1 & 82.3 & 82.5 & 81.2 & 42.3 & 83.4 & 56.1\\
    CE & BMM &  96.0 & 25.0 & 39.7 & 93.6 & 37.8 & 53.9 & 81.0 & 61.2 & 69.7 & 100.0 & 63.3 & 77.5 & 78.4 & 68.0 & 72.8 & 100.0 & 71.3 & 83.2 & 91.5 & 54.4 & 66.1 & 68.6 & 24.0 & 35.5\\
    \cline{1-26}
\end{tabular}
\end{adjustbox}
\caption{Evaluation of separation criterion and distribution modeling for known-unknown sample separation in USD on Office dataset.}
\label{res:ablation_office_model}
\end{table*}

\begin{table*}[!htbp]
\resizebox{\textwidth}{!}{$
\begin{tabular}{L|ccc|ccc|ccc|ccc|ccc|ccc|cca}
\cline{1-19}
    \multirow{2}{*}{Method} & \multicolumn{3}{c}{A $\rightarrow$ C} & \multicolumn{3}{c}{A $\rightarrow$ P} & \multicolumn{3}{c}{A $\rightarrow$ R} & \multicolumn{3}{c}{C $\rightarrow$ A} & \multicolumn{3}{c}{C $\rightarrow$ P} & \multicolumn{3}{c}{C $\rightarrow$ R} \\
    & OS* & UNK & HOS & OS* & UNK & HOS & OS* & UNK & HOS & OS* & UNK & HOS & OS* & UNK & HOS & OS* & UNK & HOS \\
    \cline{1-19}
    \rowcolor{maroon!10}
    USD (full) & 53.3 & 71.5 & 61.1 & 65.7 & 74.9 & 70.0 & 73.3 & 79.5 & 76.3 & 52.2 & 70.8 & 60.1 & 62.4 & 68.4 & 65.2 & 69.3 & 68.6 & 68.9 \\
    USD w/o $\mathcal{L}_t^{trip}$ & 52.9 & 69.9 & 60.2 & 66.4 & 75.1 & 70.4 & 73.6 & 78.9 & 76.2 & 52.0 & 70.0 & 59.3 & 62.3 & 68.5 & 65.2 & 68.0 & 67.8 & 67.9 \\ 
    USD w/o $\mathcal{L}_t^{con}$ & 50.5 & 75.6 & 60.6 & 63.3 & 77.7 & 69.8 & 69.6 & 83.1 & 75.8 & 49.4 & 74.4 & 59.3 & 57.9 & 73.8 & 64.9 & 64.3 & 72.9 & 68.3 \\
    USD w/o $\mathcal{L}_t^{IM}$ & 50.1 & 74.7 & 59.9 & 64.6 & 73.5 & 68.7 & 73.5 & 77.7 & 75.5 & 51.2 & 67.9 & 58.4 & 60.2 & 68.3 & 64.0 & 66.7 & 69.0 & 67.9 \\
    USD w/o curriculum & 47.5 & 77.1 & 58.8 & 60.8 & 79.4 & 68.9 & 69.3 & 82.5 & 75.3 & 44.7 & 79.1 & 57.1 & 57.8 & 74.6 & 65.2 & 62.2 & 73.8 & 67.5 \\ 
    USD w/o co-training & 44.0 & 80.4 & 56.8 & 58.5 & 78.4 & 67.0 & 64.1 & 78.2 & 70.5 & 43.4 & 72.3 & 54.3 & 50.5 & 71.0 & 59.0 & 51.4 & 76.1 & 61.4 \\
    \cline{1-22}

    \multirow{2}{*}{Method} & \multicolumn{3}{c}{P $\rightarrow$ A} & \multicolumn{3}{c}{P $\rightarrow$ C} & \multicolumn{3}{c}{P $\rightarrow$ R} & \multicolumn{3}{c}{R $\rightarrow$ A} & \multicolumn{3}{c}{R $\rightarrow$ C} & \multicolumn{3}{c}{R $\rightarrow$ P} & \multicolumn{3}{c}{Avg.} \\ 
    & OS* & UNK & HOS & OS* & UNK & HOS & OS* & UNK & HOS & OS* & UNK & HOS & OS* & UNK & HOS & OS* & UNK & HOS & OS* & UNK & HOS \\
    \cline{1-22}
    \rowcolor{maroon!10}
    USD (full) & 54.3 & 73.8 & 62.6 & 47.3 & 69.6 & 56.3 & 70 & 74.5 & 72.2 & 64.6 & 71.3 & 67.8 & 53.8 & 65.5 & 59.1 & 73.3 & 69.1 & 71.1 & 61.6 & 71.5 & 65.9 \\
    USD w/o $\mathcal{L}_t^{trip}$ & 51.2 & 75.9 & 61.1 & 47.6 & 70.4 & 56.8 & 69.5 & 74.2 & 71.8 & 63.9 & 69.9 & 66.8 & 51.4 & 66.9 & 58.2 & 73.5 & 67.6 & 70.4 & 61.0 & 71.3 & 65.4 \\
    USD w/o $\mathcal{L}_t^{con}$ & 49.4 & 78.0 & 60.5 & 44.9 & 71.8 & 55.3 & 66.0 & 78.4 & 71.7 & 60.3 & 75.3 & 67.0 & 50.1 & 70.1 & 58.4 & 70.6 & 73.9 & 72.2 & 58.0 & 75.4 & 65.3 \\
    USD w/o $\mathcal{L}_t^{IM}$ & 50.6 & 75.4 & 60.6 & 45.5 & 68.1 & 54.5 & 68.7 & 74.1 & 71.3 & 63.2 & 72.8 & 67.7 & 49.7 & 66.5 & 56.9 & 73.1 & 67.3 & 70.1 & 59.8 & 71.3 & 64.6 \\
    USD w/o curriculum & 46.1 & 80.7 & 58.6 & 40.4 & 74.5 & 52.4 & 64.7 & 78.3 & 70.9 & 57.9 & 77.7 & 66.4 & 48.0 & 73.3 & 58.0 & 69.0 & 73.4 & 71.1 & 55.7 & 77.0 & 64.2 \\ 
    USD w/o co-training & 48.3 & 78.9 & 59.9 & 38.5 & 71.8 & 50.1 & 51.6 & 79.2 & 62.5 & 53.9 & 76.5 & 63.2 & 46.6 & 77.6 & 58.2 & 60.7 & 80.5 & 69.2 & 51.0 & 76.7 & 61.0 \\ 
    \cline{1-22}
\end{tabular}
$}
\caption{Ablation study on the objective function, and co-training for USD on Office-Home dataset.}
\label{res:ablation_loss_officehome}
\end{table*}

\begin{table*}[!htbp]
\resizebox{1\textwidth}{!}{$
\begin{tabular}{L|ccc|ccc|ccc|ccc|ccc|ccc|cca}
\cline{1-19}
    \multirow{2}{*}{Pseudolabel} & \multicolumn{3}{c}{A $\rightarrow$ C} & \multicolumn{3}{c}{A $\rightarrow$ P} & \multicolumn{3}{c}{A $\rightarrow$ R} & \multicolumn{3}{c}{C $\rightarrow$ A} & \multicolumn{3}{c}{C $\rightarrow$ P} & \multicolumn{3}{c}{C $\rightarrow$ R} \\
    & OS* & UNK & HOS & OS* & UNK & HOS & OS* & UNK & HOS & OS* & UNK & HOS & OS* & UNK & HOS & OS* & UNK & HOS \\
    \cline{1-19}
    \rowcolor{maroon!10}
    Ensemble & 53.3 & 71.5 & 61.1 & 65.7 & 74.9 & 70.0 & 73.3 & 79.5 & 76.3 & 52.2 & 70.8 & 60.1 & 62.4 & 68.4 & 65.2 & 69.3 & 68.6 & 68.9 \\
    Clustering & 50.8 & 73.5 & 60.1 & 67.0 & 73.2 & 69.9 & 74.8 & 75.8 & 75.3 & 54.3 & 67.0 & 60.0 & 61.5 & 66.9 & 64.1 & 67.2 & 66.1 & 66.7 \\
    Student Predictions & 50.7 & 74.1 & 60.2 & 65.9	& 73.6 & 69.5 & 74.7 & 78.5 & 76.5 & 51.3 & 68.2 & 58.6 & 61.7 & 67.3 & 64.4 & 67.2 & 69.5 & 68.3 \\
    \cline{1-22}
    \multirow{2}{*}{Pseudolabel} & \multicolumn{3}{c}{P $\rightarrow$ A} & \multicolumn{3}{c}{P $\rightarrow$ C} & \multicolumn{3}{c}{P $\rightarrow$ R} & \multicolumn{3}{c}{R $\rightarrow$ A} & \multicolumn{3}{c}{R $\rightarrow$ C} & \multicolumn{3}{c}{R $\rightarrow$ P} & \multicolumn{3}{c}{Avg.} \\ 
    & OS* & UNK & HOS & OS* & UNK & HOS & OS* & UNK & HOS & OS* & UNK & HOS & OS* & UNK & HOS & OS* & UNK & HOS & OS* & UNK & HOS \\
    \cline{1-22}
    \rowcolor{maroon!10}
    Ensemble & 54.3 & 73.8 & 62.6 & 47.3 & 69.6 & 56.3 & 70.0 & 74.5 & 72.2 & 64.6 & 71.3 & 67.8 & 53.8 & 65.5 & 59.1 & 73.3 & 69.1 & 71.1 & 61.6 & 71.5 & 65.9 \\
    Clustering & 53.4 & 73.6 & 61.9 & 48.6 & 69.0 & 57.1 & 71.0 & 71.8 & 71.4 & 63.4 & 72.2 & 67.5 & 52.2 & 68.1 & 59.1 & 70.0 & 68.3 & 69.2 & 61.2 & 70.5 & 65.2\\
    Student Predictions & 53.0 & 73.2 & 61.5 & 46.6 & 71.3 & 56.4 & 68.9 & 73.8 & 71.3 & 61.8 & 73.0 & 66.9 & 53.3 & 65.8 & 58.9 & 72.7 & 70.6 & 71.6 & 60.6 & 71.6 & 65.3 \\
    \cline{1-22}
\end{tabular}
$}
\caption{Ablation study on the pseudolabeling scheme for USD on Office-Home dataset.}
\label{res:ablation_pseudo_officehome}
\end{table*}

A detailed ablation study was performed on the known-unknown sample selection criterion and on the modeling of the criterion distribution. The results of the ablation study on both Office-31 and VisDA-C are given in Table \ref{res:ablation_office_model}. 
USD uses JSD as the known-unknown sample splitting criterion, while entropy has been extensively used in existing OSDA methods (SHOT, AaD, UADAL etc.) for this purpose. 
In addition, cross-entropy (CE) loss is a popular criterion for separating clean-noisy samples for noisy label learning (NLL) algorithms \cite{arazo2019unsupervised, dividemix, Li_2022_CVPR}. 
We evaluate all three criteria to find the best performing one. 
The criterion distribution can be modelled by either Gaussian Mixture Model (GMM) or a Beta Mixture Model (BMM). 
UADAL models sample entropy distribution using BMM to distinguish between known and unknown samples.
Our results in Table \ref{res:ablation_office_model} show that modeling the distribution of the JSD with a GMM outperforms all of the other combinations for unknown sample discovery.

\begin{figure}
  \centering
    \includegraphics[width=.3\textwidth]{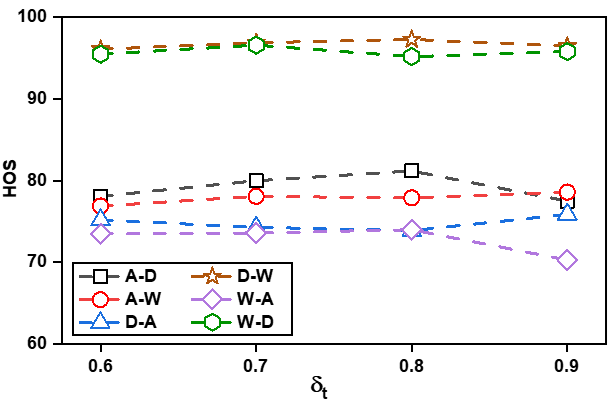}
    \caption{Impact of JSD threshold $\delta_t$ on HOS for Office dataset.}
    \label{fig:jsd_threshold}
\end{figure}

The effect of the JSD threshold $\delta_t$ for known-unknown separation on the final HOS is shown in Figure \ref{fig:jsd_threshold}. 
The performance is relatively uniform, which suggests robustness of adaptation to the hyperparameter $\delta_t$. 
Nonetheless, if the threshold is set too high (such as $0.9$), too few samples may be denoted as known samples, and this could lead to inferior performance. 

Table \ref{res:ablation_loss_officehome} shows the impact of different components of our objective function and the effects of our teacher-student co-training scheme on the final adaptation performance for Office-Home.
It is evident that each of our losses ($\mathcal{L}_{t}^{trip}, \mathcal{L}_{t}^{con}, \mathcal{L}_{t}^{IM}$) contributes to the adaptation, and leaving out any one of them hurts performance.
We observe that curriculum guidance considerably benefits adaptation and the final average HOS increases by $>1.5\%$ (from $64.2\%$ to $65.9\%$) when such guidance is included.
Notably, without curriculum, adaptation to the known classes is impacted drastically (OS* falls by $\sim6\%$), signalling that progressively learning the known class subspace first and then the unknown class subspace later is the superior strategy.

\begin{figure}
\begin{minipage}[b]{1.0\linewidth}
  \centering
    \includegraphics[width=1\textwidth]{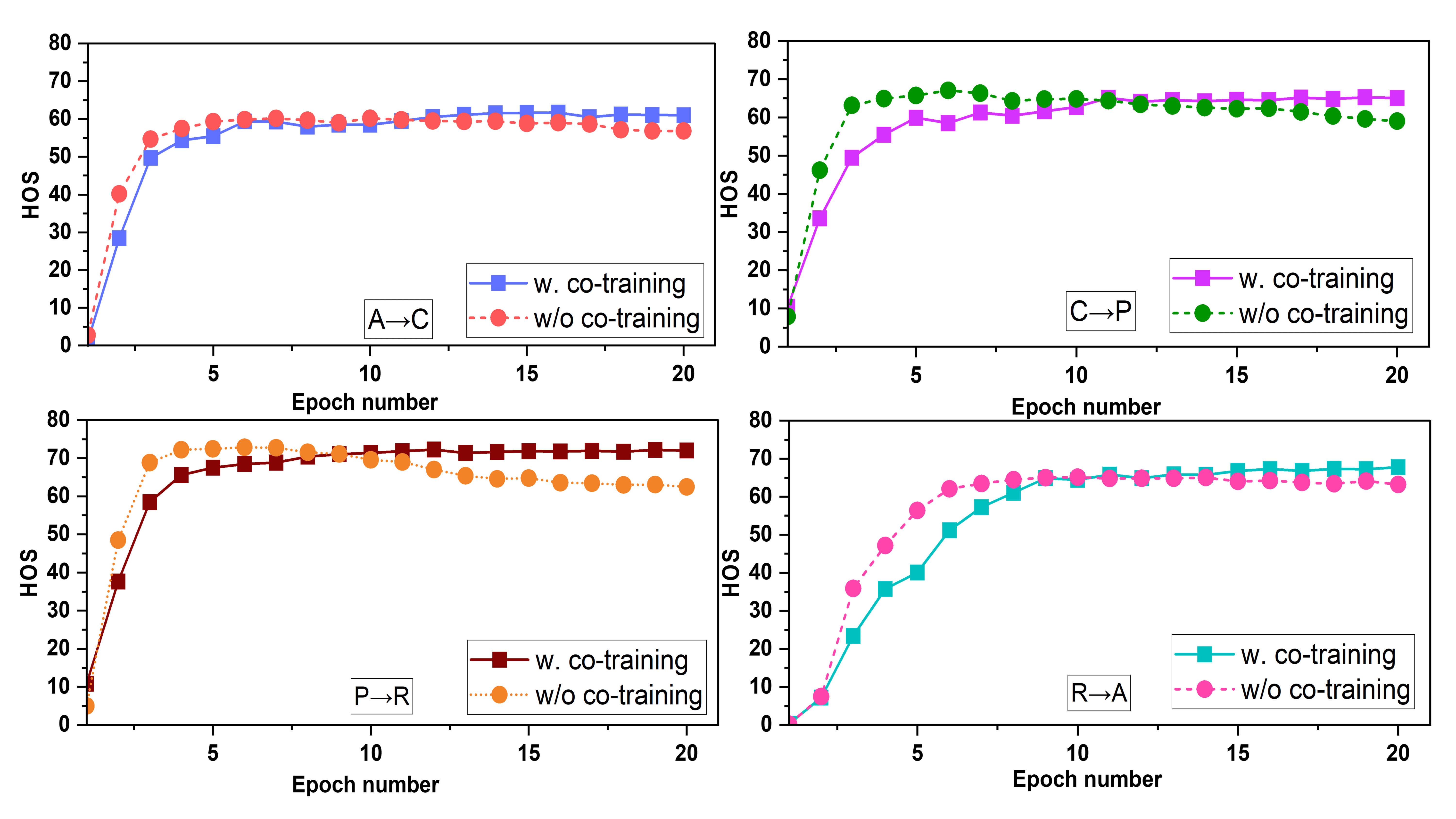}
\end{minipage}
    \caption{Impact of co-training on reducing error accumulation during adaptation on Office-Home dataset.}
    \label{fig:co-training}
\end{figure}

The final row in Table \ref{res:ablation_loss_officehome} presents results in the absence of the teacher network, where the student network conducts the known-unknown sample separation for itself. 
Both the weakly and strongly augmented samples are fed through the student network, and losses $\mathcal{L}_{t}^{trip}$, $\mathcal{L}_{t}^{con}$ are calculated over the student model outputs between the weak and strong augmentations. 
Empirical results clearly show that co-training in a teacher-student framework is pivotal for mitigating the effect of any imperfect known-unknown separation and average HOS over the 12 domain pairs in Office-Home decreases by $\sim5\%$ when the teacher network is removed.
As seen in Figure \ref{fig:co-training}, in the absence of co-training, the student model adapts faster, but its performance drops from its peak during the course of adaptation due to error accumulation.
In contrast, adaptation with co-training is slightly slower but maintains its peak performance.

The effect of the pseudolabeling scheme on the adaptation performance for Office-Home is shown in Table \ref{res:ablation_pseudo_officehome}. 
SHOT and AaD use a self-supervised clustering process built on DeepCluster \cite{caron2018deep} to get pseudolabels for the known samples.
We see that such clustering is not better than taking the hard predictions from the student model as pseudolabels. 
In open set settings, the unknown samples can drift the known class centroids, leading to faulty clusters.
Our multi-view augmentation ensembled pseudolabeling strategy outperforms both pseudolabeling from clustering or direct student predictions.

\section{Conclusion}
\label{sec:conclusion}

We present Unknown Sample Discovery as a teacher-student co-training framework that conducts SF-OSDA by splitting the target data into known and unknown subsets based on the JSD criterion modeled with a 2-component Gaussian mixture model. 
Co-training regularization greatly mitigates error accumulation, while curriculum guidance progressively adapts the target model to effectively learn both the known and unknown target feature spaces.
Empirically, USD outperforms existing SF-OSDA methods and is comparable to non-source-free OSDA techniques. 
{
    \small
    \bibliographystyle{ieeenat_fullname}
    \bibliography{main}
}


\end{document}